\documentclass[runningheads]{llncs}

\usepackage{eccv}

\usepackage{eccvabbrv}

\usepackage{graphicx}
\usepackage{booktabs}

\usepackage[accsupp]{axessibility}  

\usepackage{hyperref}

\usepackage{orcidlink}

\usepackage{marvosym}
\usepackage[inkscapelatex=false]{svg}
\usepackage{multirow}
\newcommand{\tbl}[1]{Tab. \ref{#1}}
\newcommand{\fig}[1]{Fig. \ref{#1}}
\newcommand{\eq}[1]{Eq. (\ref{#1})}

\begin{document}

\title{MobileIQA: Exploiting Mobile-level Diverse Opinion Network For No-Reference Image Quality Assessment Using Knowledge Distillation} 

\titlerunning{ Mobile-level Diverse Opinion NR-IQA Using Knowledge Distillation}

\author{Zewen Chen\inst{1,2}\orcidlink{0000-0001-7791-0959}
 \and
 Sunhan Xu\inst{3} \and
 Yun Zeng\inst{4} \orcidlink{0009-0004-5934-1528} \and
 Haochen Guo \inst{5} \and
 Jian Guo \inst{3} \and
 Shuai Liu \inst{3} \and
Juan Wang\inst{1} \orcidlink{0000-0002-3848-9433}\and
Bing  Li\inst{1,6}\textsuperscript{\Letter}\orcidlink{0000-0001-6114-1411} \and
Weiming Hu\inst{1,2,7}\orcidlink{0000-0001-9237-8825}  \and
Dehua Liu\inst{8} \and
Hesong Li\inst{8}
}

\authorrunning{Z.Chen et al.}

\institute{
State Key Laboratory of Multimodal Artificial Intelligence Systems, CASIA \and
School of Artificial Intelligence, University of Chinese Academy of Sciences 
\makebox[0.5\textwidth][c]{\and Beijing Union University \and China University of Petroleum \and Hebei University} \and
PeopleAl Inc. Beijing, China \and
School of Information Science and Technology, ShanghaiTech University \and
Shanghai Transsion Information Technology Limited  \\
\email{chenzewen2022@ia.ac.cn}, \email{20221081210206@buu.edu.cn}, 
\email{\{cup\_zy1, hcguo\_hbu\}@163.com}, 
\email{1418319765@qq.com}, 
\email{20231081210210@buu.edu.cn}, 
\email{jun\_wang@ia.ac.cn}, 
\email{\{bli, wmhu\}@nlpr.ia.ac.cn}, \\
\email{\{dehua.liu, hesong.li\}@transsion.com}
}

\maketitle

\begin{abstract}
With the rising demand for high-resolution (HR) images, No-Reference Image Quality Assessment (NR-IQA) gains more attention, as it can ecaluate image quality in real-time on mobile devices and enhance user experience. However, existing NR-IQA methods often resize or crop the HR images into small resolution, which leads to a loss of important details. And most of them are of high computational complexity, which hinders their application on mobile devices due to limited computational resources. To address these challenges, we propose MobileIQA, a novel approach that utilizes lightweight backbones to efficiently assess image quality while preserving image details through high-resolution input. MobileIQA employs the proposed multi-view attention learning (MAL) module to capture diverse opinions, simulating subjective opinions provided by different annotators during the dataset annotation process. The model uses a teacher model to guide the learning of a student model through knowledge distillation. This method significantly reduces computational complexity while maintaining high performance. Experiments demonstrate that MobileIQA outperforms novel IQA methods on evaluation metrics and computational efficiency. The code is available at \url{https://github.com/chencn2020/MobileIQA}.
\end{abstract}

\section{Introduction}
\label{sec:intro}

Image quality assessment (IQA) is a long-standing research in image processing fields.
According to the availability of reference images, IQA can be categorized into three types: full-reference IQA (FR-IQA), reduced-reference IQA (RR-IQA) and no-reference IQA (NR-IQA).
Among these types, NR-IQA has gained more attention since it removes the dependence on reference images, which are unavailable in many real-world applications.

\begin{figure}[htbp]
  \centering
  \includegraphics[width=\columnwidth]{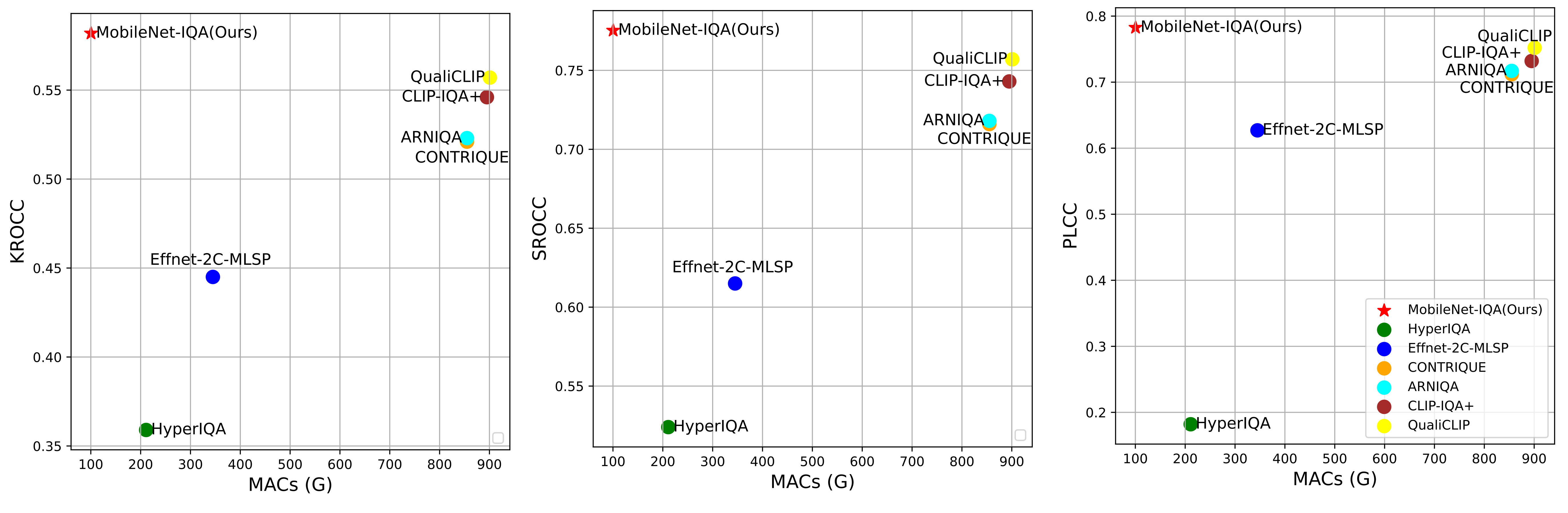}
  \caption{Comparison among SOTA IQA methods on UHD-IQA \cite{hosu2024uhd} validation set in terms of KROCC, SROCC, PLCC and MACs.}
  \label{fig:intro}
\end{figure}

With the development of mobile imaging technology, capturing high-resolution (HR) images (such as 4K) using mobile devices, such as cameras and smartphones, has become increasingly popular. The higher the quality of these images is, the better the user experience will be. Therefore, evaluating the quality of HR images in real-time on mobile devices is crucial.

Over the past decades, numerous efforts have been adopted to NR-IQA, such as developing sophisticated networks\cite{ke2021musiq, yang2022maniqa, chen2024promptiqa}, proposing proxy tasks\cite{chen2022teacher, liu2017rankiqa, wang2023Hierarchical}, introducing Vision-Language Models (VLM) \cite{wu2023qinstruct, wu2024qbench}. Although these methods have improved the performance of IQA models in various aspects, they still encounter two major challenges when assessing the quality of HR images on mobile devices. \textbf{(1) Limited Input Resolution}: Most methods resize or crop the HR images into smaller resolution, typically $224\times224$, which represents only about 1\% of the resolution of 4K images. This process results in the loss of important image details, thereby limiting the model's generalization and performance.
\textbf{(2) High Computational Complexity}: Most of these methods employ computationally intensive backbones such as ResNet \cite{he2016deep} or vision transformer (ViT) \cite{dosovitskiy2020image}. However, the limited computational resources available on mobile devices make it challenging to efficiently run these models on such platforms. 
The two challenges significantly hinder the application of these IQA methods on mobile devices.

In this paper, we introduce MobileIQA, which achieves outstanding performance with significantly fewer multiply-accumulate operations (MACs) to tackle these challenges. 
IQA is an extremely subjective task, since different individuals perceive the quality differently, leading to variations in their quality ratings of the same image.
Therefore, the ground truth (GT)  labels of images are defined as the average of subjective scores provided by multiple human annotators, namely mean opinion score (MOS). Mimicking the human rating process, we develop a multi-view attention learning (MAL) module for the MobileIQA to implicitly learn diverse opinion features by capturing complementary contexts from various perspectives. The opinion features collected from different MALs are integrated into a comprehensive quality score, effectively facilitating more reliable quality score assessment. 

MobileIQA consists of a teacher model (MobileViT-IQA) and a student model (MobileNet-IQA), which utilize lightweight MobileViT \cite{mehta2021MobileViT} and MobileNet \cite{howard2017mobilenets} as backbones respectively. These networks with  lightweight backbones support a maximum resolution of $1907\times1231$, effectively preserving the details in HR images.  Although MobileViT-IQA outperforms MobileNet-IQA due to its global attention mechanism, it is less computational efficiency. To address this, we employ knowledge distillation, using MobileViT-IQA as the teacher network to guide the learning of MobileNet-IQA. This approach significantly reduces the computational complexity and improves the performance of MobileNet-IQA. As shown in \fig{fig:intro}, our model demonstrate excellent performance in terms of three evaluation metrics and MACs compared to the novel comparison IQA models.
Overall, our contributions are summarized as follows:

\begin{enumerate}
    \item We propose MobileIQA, which integrates diverse opinion features produced by our meticulously designed MAL modules, effectively enhancing the performance of the model. 
    \item We employ knowledge distillation to transfer the knowledge from the teacher network to the student network, thereby significantly reducing the computational complexity while maintaining the performance.
    \item Numerous experimental results demonstrate that our MobileNet-IQA achieves higher accuracy and computational efficiency, significantly outperforming many advanced methods.
    
\end{enumerate}

\section{Related Works}

Due to the remarkable progress in vision applications, considerable attention has been focused on elevating the performance of IQA. As a pioneer, \cite{kang2014convolutional} design a convolutional neural network (CNN) for IQA to extract image features. Then they extend this work to a multi-task CNN \cite{kang2015simultaneous}.
However, insufficient training samples limit effective learning of CNNs-based models. 
For this reason, some methods \cite{su2020blindly,qin2023data,zhang2018blind} employ pre-trained networks, such as ResNet \cite{he2016deep} and ViT \cite{dosovitskiy2020image}, as feature extractors.
However, recent research \cite{zhu2020metaiqa,chen2022teacher} point out that these popular networks pre-trained for high-level tasks are not suitable for IQA.
Therefore, some works pre-train models on related pretext tasks, \emph{e.g.}, image restoration \cite{lin2018hallucinated,ma2021blind}, quality ranking \cite{liu2017rankiqa,ma2017dipiq}, and contrastive learning \cite{zhao2023quality,madhusudana2022image}. 
Some other methods enhance the IQA performance by introducing auxiliary information. For instance, Wang et al. and Saha et al.
\cite{wang2023exploring,saha2023re} integrate textual information into the IQA.
Zhang et al. \cite{zhang2023blind} explore the relationship among multiple tasks, namely the IQA, scene classification and distortion classification.
Additionally, many methods utilize the idea of ensemble learning to aggregate IQA-related knowledge for more effective learning. 
\cite{ma2019blind} collect a set of existing IQA models for annotation. The annotated samples are used for training their model to learn the quality score as well as the uncertainty.
Some methods \cite{wang2023Hierarchical,zhang2020learning,zhang2021uncertainty} propose a novel multi-dataset training strategy. 
The IQA task is also approached as a quality ranking problem. Gao et al.\cite{2013Learning} utilize cross-entropy loss to measure the discrepancy between predicted quality rankings and GT binary labels for each image pair. Liu et al.\cite{liu2017rankiqa} use hinge loss to define the optimization objective for quality ranking learning, while Ma et al.\cite{ma2017dipiq} apply learning-to-rank algorithms like RankNet\cite{2005Learning} and ListNet\cite{cao2007learning} to train IQA models on numerous image pairs. 

Although existing methods have improved IQA performance by addressing various aspects of the model, they take the traditional computer vision resolutions, such as $224\times224$ or $256\times256$ as the input images, which limits the adaptability to the HR IQA task. 
Additionally, most of them utilize computationally intensive backbones like ResNet or ViT, making it challenge to be applied on resource-constrained mobile devices. To address this, we propose MobileIQA, a mobile-level IQA model based on diverse opinion and knowledge distillation. By leveraging lightweight backbones, and employing knowledge distillation, our model significantly reduces computational complexity while maintaining model performance.

\section{Proposed Method}

\subsection{Model Design}

\begin{figure}[htbp]
    \centering
    \includegraphics[width=\textwidth]{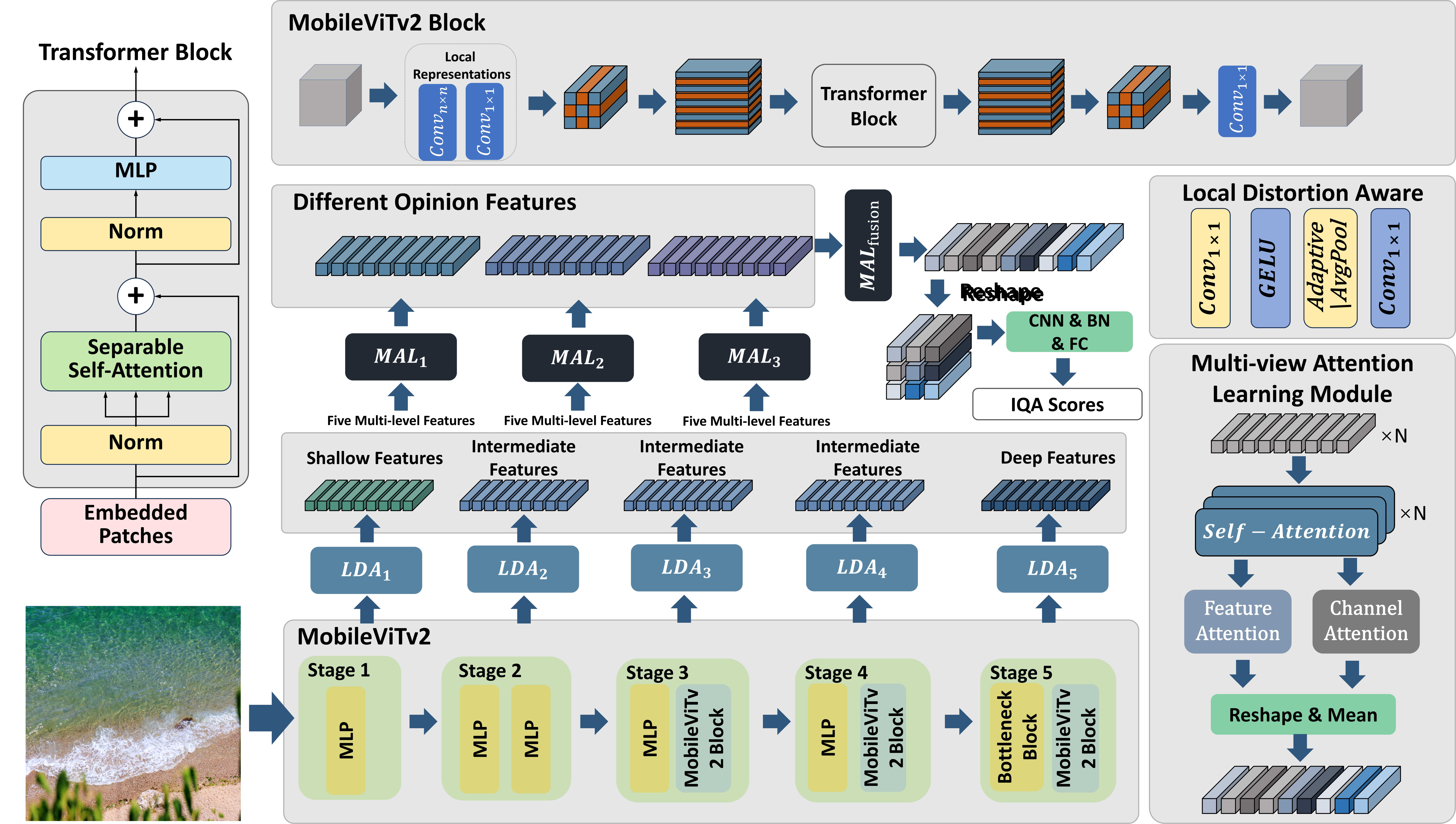}
    \caption{Framework of the teacher model (MobileViT-IQA). The student model (MobileNet-IQA) shares the same framework, but takes MobileNet as backbone.}
    \label{fig:teacher}
\end{figure}

In this work, we present a novel network called MobileIQA, which uses teacher-student distillation \cite{hu2023teacher} as the training technique. Both of the teacher and student model take lightweight backbones for feature extraction and collects various opinions by capturing diverse attention contexts to make a comprehensive decision on the image quality score. \fig{fig:teacher} shows the teacher network (MobileViT-IQA) architecture in the MobileIQA, which mainly consists of four parts: (1) A pre-trained MobileViT employed for multi-level feature perception; (2) Local distortion aware (LDA) modules used for unifying multi-level feature dimensions; (3) Multi-view attention learning (MAL) modules proposed for opinion collection; (4) An image quality score regression module designed for quality estimation. The architecture of MobileNet-IQA is similar to the MobileViT-IQA, but uses the MobileNet as the backbone. In the following, we introduce the MobileViT-IQA in detail.

\subsubsection{(A) Multi-level Feature Perception.}
The blocks in MobileViT replace local processing in traditional CNNs with global processing via transformers, integrating characteristics of both CNNs and ViTs. This architecture enables the MobileViT to learn representations more efficiently.
Given an image $I \in \mathbb{R}^{3\times H \times W }$, we extract the features from the MobileVit. Many existing work proves that the mutli-layer features are useful for the IQA task\cite{chen2022teacher, chen2024promptiqa, wang2023Hierarchical, su2020blindly,hosu2019effective}. Thus, we extract multi-level features from the five stages in MobileViT, denoted as \( f_j \in \mathbb{R}^{C_j \times H_j \times W_j} \), where \( C_j \), \( H_j \), and \( W_j \) represent the dimension of the feature map at the \( j \)-th stage and \( 1 \leq j \leq 5 \).

\subsubsection{(B) Local Distortion Aware Module. } The Local Distortion Aware (LDA) module serves two key functions: (1) It extracts local features using a CNN with a small receptive field; (2) It standardizes the dimensions of these features using an adaptive pooling operation. Specifically, for an input feature \( f_i \in \mathbb{R}^{(C_j \times H_j \times W_j)} \), a \( 1 \times 1 \) CNN is applied to double the channel dimensions to \( 2 \times C_j \). After GELU activation, the adaptive pooling operation reshapes the feature into \( f_i \in \mathbb{R}^{(2C_j \times D \times N)} \), where \( D \) and \( N \) denote the dimensions. Another \( 1 \times 1 \) CNN is used to reduce the channel dimensions back to \( C_j \), producing aware features  \( f_i \in \mathbb{R}^{C_i \times D \times N} \) for the \( i \)-th stage.

\subsubsection{(C) Multi-view Attention Learning Module.}
The critical part of the MobileIQA is the multi-view attention learning (MAL) module. The motivation behind it is that individuals often have diverse subjective perceptions and regions of interest when viewing the same image. To this end, we employ multiple MALs to learn attentions from different viewpoints.
Each MAL is initialized with different weights and updated independently to encourage diversity and avoid redundant output features. The number of MALs can be flexibly set as a hyper-parameter. In this work, we set it to 3 and we show in our results its effect on the performance of our model.

As shown in \fig{fig:teacher}, the MAL starts from $N$ self-attentions (SAs), each of which is responsible to process a basic feature $\mathbf{f}_j$ ($1\leq j \leq N$). 
The outputs of all the SAs are concatenated, forming a multi-level aware feature $\mathbf{F}\in \mathbb{R}^{C\times D \times N}$. Then $\mathbf{F}$ passes through two branches, \emph{i.e.}, a feature-wise SA branch and a channel-wise SA branch, which apply a SA across spatial and channel dimensions, respectively, to capture complementary non-local contexts and generate multi-view attention maps.
In particular, for the channel-wise SA, the feature $\mathbf{F}$ is first reshaped and permuted to convert the size from $C\times D \times N$ to $D\times (C \times N)$. After the SA, the output feature is permuted and reshaped back to the original size $C\times D \times N$. Subsequently, the outputs of the two branches are added and average pooled, generating an opinion feature.
The design of the two branches has two key advantages. 
First, implementing the SA in different dimensions promotes diverse attention learning, yielding complementary information. Second, contextualized long-range relationships are aggregated, benefiting global quality perception.

In MobileIQA, there are four MALs in total. Three of them independently extract opinion features from the five-level features captured from the LDAs, representing the perspectives of different annotators during data annotation. The fourth MAL fuses these three opinion features into a final quality feature.

 \subsubsection{(D) Image Quality Score Regression.}
Assuming that $M$ opinion features are generated from $M$ MALs employed in the MobileIQA. To derive a global quality score from the collected opinion features, we utilize an additional MAL. The MAL integrates diverse contextual perspectives, resulting in a comprehensive opinion feature that captures essential information. This feature is then processed through two CNN layers with kernel sizes of $1 \times 1$ and $3 \times 3$ to reduce the number of channels, followed by two fully connected layers that transform the feature size from 128 to 64 and from 64 to 1. Finally, we obtain a predicted quality score.

\subsection{Knowledge Distillation}

\begin{figure*}[htbp]
    \centering
    \includegraphics[width=\textwidth]{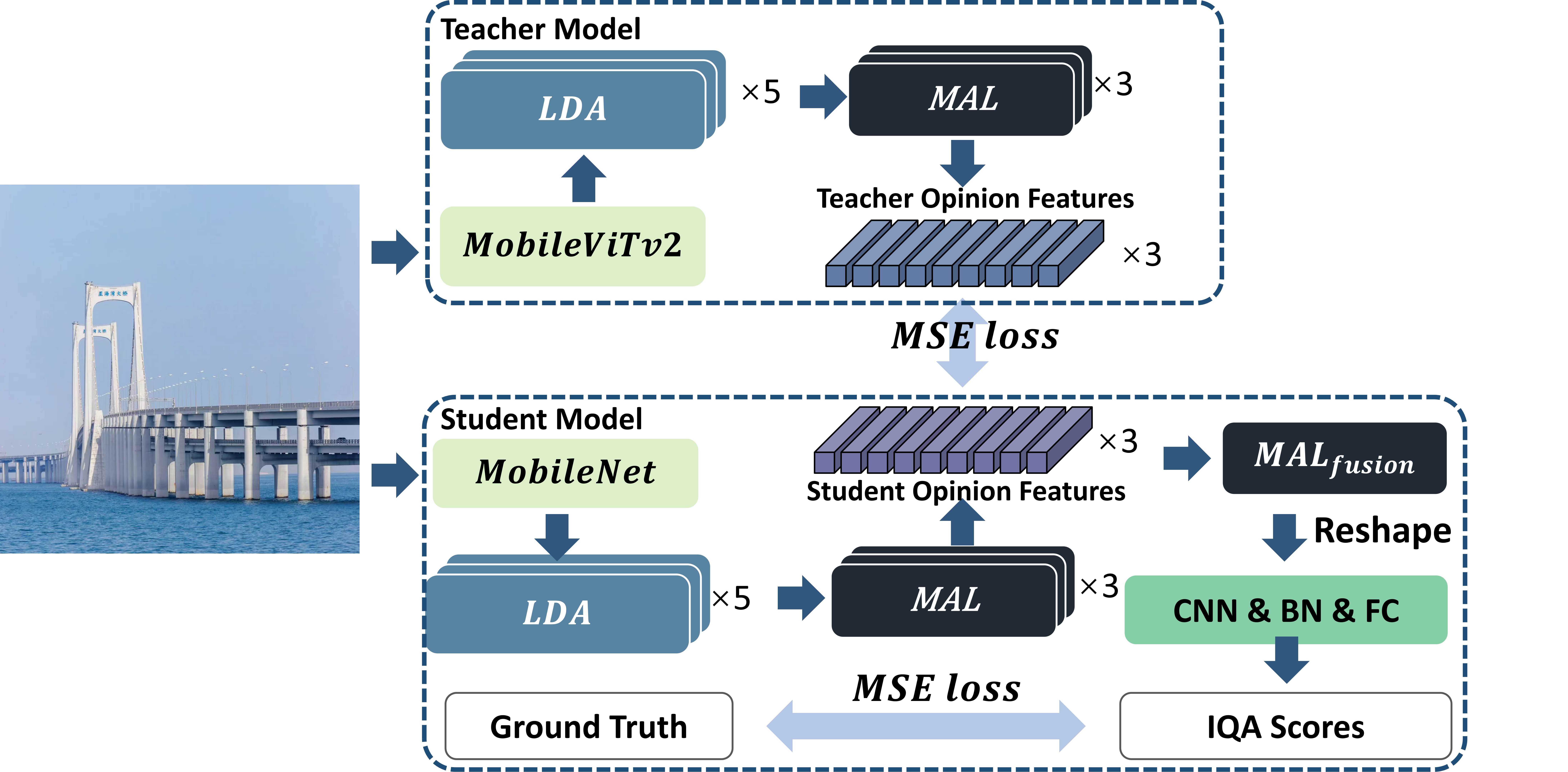}
    \caption{Knowledge distillation process. MSE loss is used to minimize the discrepancy between the Student Opinion Features and the Teacher Opinion Features.}
    \label{fig:distill}
\end{figure*}

Despite the superior performance of MobileViT-IQA, its computational complexity still poses a burden on mobile devices. In contrast, MobileNet-IQA requires less computation but does not match the performance of MobileViT-IQA. To address this issue, we design a distillation process, as illustrated in \fig{fig:distill}, where MobileViT-IQA serves as the teacher model, guiding the learning of the student model MobileNet-IQA. Since MobileNet-IQA and MobileViT-IQA share the same architecture except for the backbone, the distillation process is pretty easy and efficient. Considering that different MALs in MobileIQA simulate the opinions from different evaluators, we apply $MSE$ loss to minimize the discrepancy between the MAL outputs from the teacher model and the student model, thereby enabling the MALs in the student to approximate the opinions from MALs in the teacher.

Specifically, given an image $I\in\mathbb{R}^{3\times H\times W}$, the teacher and student models extract the multi-level aware features for the all five stages $f_i^T$ and $f_i^T$ respectively. These features are then processed by three MALs in both models, producing teacher opinion features (\( \mathbf{F}_i^T \)) and student opinion features (\( \mathbf{F}_i^S \)). The discrepancy between these two types of opinion features is minimized using an MSE loss, effectively allowing the teacher model to guide the student model in how to assess images. This process can be formulated as follows:
\begin{equation}
    l_d = \frac{1}{3}\sum_{i=1}^3\mathbf{MSE}(\mathbf{F}_i^T, \mathbf{F}_i^S).
\end{equation}

Meanwhile, to improve the score prediction accuracy of the student model, we additionally employ the MSE loss during the distillation process to minimize the discrepancy between the student's predicted scores and the GTs. The optimization objective  for the distillation is to minimize the following loss function:
\begin{equation}
    l = l_d + \alpha\times \mathbf{MSE}(P, G),
    \label{eq:distill}
\end{equation}
where \( P \) represents the predicted score and \( G \) the ground truth, with \( \alpha \) denoting a constant.

\section{Experiments}

\subsection{Datasets}

We train and evaluate our model on UHD-IQA \cite{hosu2024uhd} dataset, totally containing 6,073 HR images, where 4269 and 904 are used for training and validating, respectively. The organizers in \textbf{UHD-IQA Challenge: Pushing the Boundaries of Blind Photo Quality Assessment}~\cite{aim2024uhdbpqa} held by AIM 2024 Workshop~\footnote{\url{https://www.cvlai.net/aim/2024/}} use the remaining  900  inaccessible images as the test set to evaluate the performance. For training and distillation, only the training set from UHD-IQA is used, without any additional datasets.
\subsection{Evaluation Metrics}
We evaluate the performance of IQA models using five metrics: Kendall Rank Correlation Coefficient (KRCC), Spearman Rank-Order Correlation Coefficient (SRCC), Pearson Linear Correlation Coefficient (PLCC), Root Mean Square Error (RMSE) and  Mean Absolute Error (MAE). SRCC and KRCC assess the monotonicity, PLCC measures the linearity of the model's predictions, RMSE and MAE indicates prediction accuracy. An effective IQA model should aim for KRCC, SRCC, and PLCC values approaching 1, while minimizing RMSE and MAE values to 0.

\subsection{Implementation Details}
\label{subsec:Implementation}

We take the pre-trained mobilevitv2\_200 and mobilenetv3\_large\_100 as the backbone of the MobileViT-IQA and MobileNet-IQA. 
If not explicitly specified, the number of the MAL is set to 3 and the input images are resized into $1907\times1231$, which is the maximum training resolution that our hardware can support, during training and testing.
We set the constant $\alpha = 2$ in the \eq{eq:distill}.
We use the Adam optimizer with a learning rate of $1 \times 10^{-5}$ and a weight decay of $1 \times 10^{-5}$.  The learning rate is adjusted using the Cosine Annealing for every 50 epochs. We train the teacher model for 100 epochs with a batch size of 4 and the student model for 300 epochs with a batch size of on 8 on one NVIDIA RTXA800.

\subsection{Comparisons With State-of-the-Arts}

We compare our model with 6 advanced IQA models, namely 
HyperIQA \cite{su2020blindly},  
Effnet-2C-MLSP \cite{wiedemann2023KonxCrossresolutionImageb},	 
CONTRIQUE \cite{madhusudana2022image},	 
ARNIQA \cite{agnolucci2024arniqa},	 
CLIP-IQA+ \cite{wang2023exploring} and
QualiCLIP \cite{agnolucci2024qualityaware}. 
Following \cite{hosu2024uhd}, the computational efficiency of all these models is measured by the number of MACs required for a forward pass with the same input image size of $3840 \times 2160$. 

The results on the validation and test set in the UHD-IQA datasets are shown in \tbl{tab:valid} and \tbl{tab:test}. 
The proposed MobileNet-IQA significantly outperforms the comparison methods in terms of both performance and computational efficiency.
Particularly, compared to the comparison state-of-the-art (SOTA) models, namely QualiCLIP, our MobileNet-IQA model demonstrates significant improvements in key metrics. On the validation and  set, it achieves increases of 4.49\% and 4.91\% in KRCC, 4.12\% and 4.28\% in PLCC, 44.30\% and 20.48\% in RMSE, 46.88\% and 30.30\% in MAE, and 2.38\% and 2.34\% in SRCC,  while reducing computational complexity by 88.90\%. 
Compared to HyperIQA, which has MACs closer to ours, MobileNet-IQA significantly outperforms in all five metrics, with improvements ranging from 38.18\% to 330.22\% on the validation set and 34.29\% to 633.98\% on the test set. These results highlight the clear advantages of our proposed method over most existing IQA models.

It is worth noting that through knowledge distillation, the performance of MobileNet-IQA across five metrics is only slightly lower than that of the teacher model (MobileViT-IQA), with a maximum performance drop of just 0.003,  while significantly enhancing computational efficiency by approximately 91.66\%.
This clearly demonstrates that our designed network architecture and knowledge distillation approach significantly improve computational efficiency while maintaining the performance of the student network.

We also list the results of AIM 2024 UHD-IQA Challenge in \tbl{tab:competition}.  It shows that our model achieves the fourth place, which further demonstrates the effectiveness of our model.

\begin{table}[]
    \caption{Evaluation of the performance of the baselines on the validation set. $\uparrow$ means that higher values are better, $\downarrow$ means that lower values are better. Best and second-best results are highlighted in bold and underlined, respectively.}
    \centering
    \begin{tabular}{lcccccc}
        \toprule
        Method & KRCC$\:\uparrow$ & PLCC$\:\uparrow$ & RMSE$\:\downarrow$ & MAE$\:\downarrow$ & SRCC$\:\uparrow$ & MACs (G)$\:\downarrow$ \\ \midrule
        HyperIQA \cite{su2020blindly} & 0.359 & 0.182 & 0.087 & 0.055 & 0.524 & \underline{211} \\		
        Effnet-2C-MLSP \cite{wiedemann2023KonxCrossresolutionImageb} & 0.445 & 0.627 & 0.060 & 0.050 & 0.615 & 345 \\
        CONTRIQUE \cite{madhusudana2022image} & 0.521 & 0.712 & 0.049 & 0.038 & 0.716 & 855 \\
        ARNIQA \cite{agnolucci2024arniqa} & 0.523 & 0.717 & 0.050 & 0.039 & 0.718 & 855 \\
        CLIP-IQA+ \cite{wang2023exploring} & 0.546 & 0.732 & 0.108 & 0.087 & 0.743 & 895 \\
        QualiCLIP \cite{agnolucci2024qualityaware} & 0.557 & 0.752 & 0.079 & 0.064 & 0.757 & 901 \\ \hline
        MobileViT-IQA(Teacher) & \textbf{0.585} & \textbf{0.784} & \textbf{0.043} & \textbf{0.034} & \textbf{0.777} & 1199 \\
        MobileNet-IQA(Student) & \underline{0.582} & \underline{0.783} & \underline{0.044} & \textbf{0.034} & \underline{0.775} & \textbf{100} \\ \bottomrule
    \end{tabular}
    \label{tab:valid}
\end{table}

\begin{table}[]
    \caption{Evaluation of the performance of the baselines on the test set. Best and second-best results are highlighted in bold and underlined, respectively.}
    \centering
    \begin{tabular}{lcccccc}
        \toprule
        Method & KRCC$\:\uparrow$ & PLCC$\:\uparrow$ & RMSE$\:\downarrow$ & MAE$\:\downarrow$ & SRCC$\:\uparrow$ & MACs (G)$\:\downarrow$ \\ \midrule
        HyperIQA \cite{su2020blindly} & 0.389 & 0.103 & 0.118 & 0.070 & 0.553 & \underline{211} \\
        Effnet-2C-MLSP \cite{wiedemann2023KonxCrossresolutionImageb} & 0.491 & 0.641 & 0.074 & 0.059 & 0.675 & 345 \\
        CONTRIQUE \cite{madhusudana2022image} & 0.532 & 0.678 & \underline{0.073} & \underline{0.052} & 0.732 & 855 \\
        ARNIQA \cite{agnolucci2024arniqa} & 0.544 & 0.694 & 0.074 & \underline{0.052} & 0.739 & 855 \\
        CLIP-IQA+ \cite{wang2023exploring} & 0.551 & 0.709 & 0.111 & 0.089 & 0.747 & 895 \\
        QualiCLIP \cite{agnolucci2024qualityaware} & \underline{0.570} & \underline{0.725} & 0.083 & 0.066 & \underline{0.770} & 901 \\ \hline
        MobileNet-IQA(Student) & \textbf{0.598} & \textbf{0.756} & \textbf{0.066} & \textbf{0.046} & \textbf{0.788} & \textbf{100} \\\bottomrule
        \end{tabular}
    \label{tab:test}
\end{table}

\begin{table}[]
\caption{The results on the private test set of AIM 2024 UHD-IQA Challenge\protect\footnotemark.}
\label{tab:competition}
    \centering
\begin{tabular}{cccccc}
\toprule
Models & MAE$\:\downarrow$ & RMSE$\:\downarrow$ & PLCC$\:\uparrow$ & SRCC$\:\uparrow$ & KRCC$\:\uparrow$ \\ \midrule
SJTU\_MMLab & 0.042 & 0.061 & 0.798 & 0.846 & 0.657 \\
CIPLAB & 0.044 & 0.064 & 0.800 & 0.835 & 0.642 \\
ZX\_AIE\_Vector\_MACs\_compute\_file & 0.044 & 0.062 & 0.768 & 0.795 & 0.605 \\
\textbf{I$^2$Group (Ours)} & \textbf{0.046} & \textbf{0.066} & \textbf{0.756} & \textbf{0.788} & \textbf{0.598} \\
Baseline & 0.049 & 0.070 & 0.722 & 0.772 & 0.581 \\
Dominator & 0.052 & 0.072 & 0.712 & 0.731 & 0.539 \\ 
ICL & 0.115 & 0.136 & 0.521 & 0.517 & 0.361 \\\bottomrule
\end{tabular}
\end{table}
\footnotetext{Results exceeding the competition's computational limits are excluded.}

\subsection{ Discussion about the Number of the MAL}
To explore the effect of the MAL's number $M$ on the performance of our model, 
we re-train the MobileViT-IQA using different settings of $M$ (1, 2 and 3).
The results on the validation set are illustrated in \tbl{tab:mal_num}. We can see that with the increase of the number of MALs, MobileViT-IQA consistently demonstrates an improved performance. This indicates that incorporating more MALs can benefit the performance, since more complementary contexts are learned. 
Additionally, we find that the discrepancy metrics (RMSE and MAE) remain unchanged, while the consistency (KRCC, PLCC and SRCC) show significant variation. We speculate that the additional complementary contexts provided by different MALs contribute to a more stable prediction of quality scores, leading to more reliable ranking and correlation rather than changes in absolute scores.

\begin{table}[htbp]
\centering
\caption{The impact of the MAL's number on the performance of MobileViT-IQA on validation set. The average results of KRCC, PLCC and SRCC are provided. The best results are marked in black bold.}
\label{tab:mal_num}
\begin{tabular}{c|cc|cccc}
\toprule
  MAL Num & RMSE$\:\downarrow$ & MAE$\:\downarrow$ & KRCC$\:\uparrow$ & PLCC$\:\uparrow$ & SRCC$\:\uparrow$ &  Average$\:\uparrow$ \\ \midrule
 1 & \textbf{0.043} & \textbf{0.034} & 0.575 & 0.775 & 0.767 & 0.706 \\
  2 & \textbf{0.043} & \textbf{0.034} & 0.576 & 0.780 & 0.770 & 0.709 \\
  3 & \textbf{0.043} & \textbf{0.034} & \textbf{0.585} & \textbf{0.784} & \textbf{0.777} & \textbf{0.715} \\ \bottomrule
\end{tabular}
\end{table}

\subsection{Discussion about the impact of the resolution of input images}
To investigate the impact of different input resolutions on model performance, we directly resize the original 4K resolution images ($3840 \times 2160$) into smaller sizes, namely $238\times 153$, $224\times 224$, $317\times 205$, $476\times 307$, $1271\times 820$ and $1907\times 1231$. We re-train the MobileViT-IQA with these 7 different types of resolutions. The results are summarized in \tbl{tab:resolusion}, where the ``Area Rate'' denotes the ratio of the input resolution to the 4K resolution. 

The results indicate that when the input resolution area rate is less than 1\% of the 4K resolution (such as $224 \times 224$), there is a significant drop in model performance. This degradation is due to the substantial loss of detailed information when high-resolution images are resized to low resolutions. 
As resolution increases, model performance improves significantly. Specifically, when the resolution is increased from $476 \times 307$ (1.76\%) to $1271 \times 820$ (12.57\%), performance metrics improve by 6.8\% to 20.0\%, where the resolution increases by approximately 7.13\%.
However, further increasing the resolution from $1271 \times 820$ (12.57\%) to $1907 \times 1231$ (28.30\%) results in minimal performance improvement. This could be due to the relatively small difference between this two resolutions (about 2.25\%), which may not significantly affect the model. Due to GPU computational limitations, further investigation with higher resolutions has not yet been conducted.

\begin{table}[]
\caption{The impact of the resolution of input images on the performance of MobileViT-IQA on the validation set. The average results of KRCC, PLCC and SRCC are provided. The best results are highlighted in bold.}
\label{tab:resolusion}
\begin{tabular}{cc|cc|cccc}
\toprule
Input Resolution & Area Rate & RMSE$\:\downarrow$ & MAE$\:\downarrow$ & KRCC$\:\uparrow$ & PLCC$\:\uparrow$ & SRCC$\:\uparrow$ &  Average$\:\uparrow$ \\ \midrule
 $238\times 153$ & 0.44\% & 0.058 & 0.047 & 0.316 & 0.477 & 0.458 & 0.417 \\
  $224 \times224$ & 0.60\% & 0.058 & 0.046 & 0.339 & 0.505 & 0.488 & 0.444 \\
  $317\times 205$ & 0.78\% & 0.056 & 0.045 & 0.380 & 0.555 & 0.542 & 0.493 \\
  $476\times 307$ & 1.76\% & 0.052 & 0.041 & 0.456 & 0.652 & 0.637 & 0.582 \\
  $1271\times 820 $& 12.57\% & 0.043 & \textbf{0.033} & 0.578 & 0.782 & 0.770 & 0.710 \\
  $1907\times1231$ & 28.30\% & \textbf{0.043} & 0.034 & \textbf{0.585} & \textbf{0.784} & \textbf{0.777} & \textbf{0.715} \\ \bottomrule
\end{tabular}
\end{table}

\subsection{ Visualization Analysis on the MAL}
\begin{figure}[]
  \centering
  \includegraphics[width=\columnwidth]{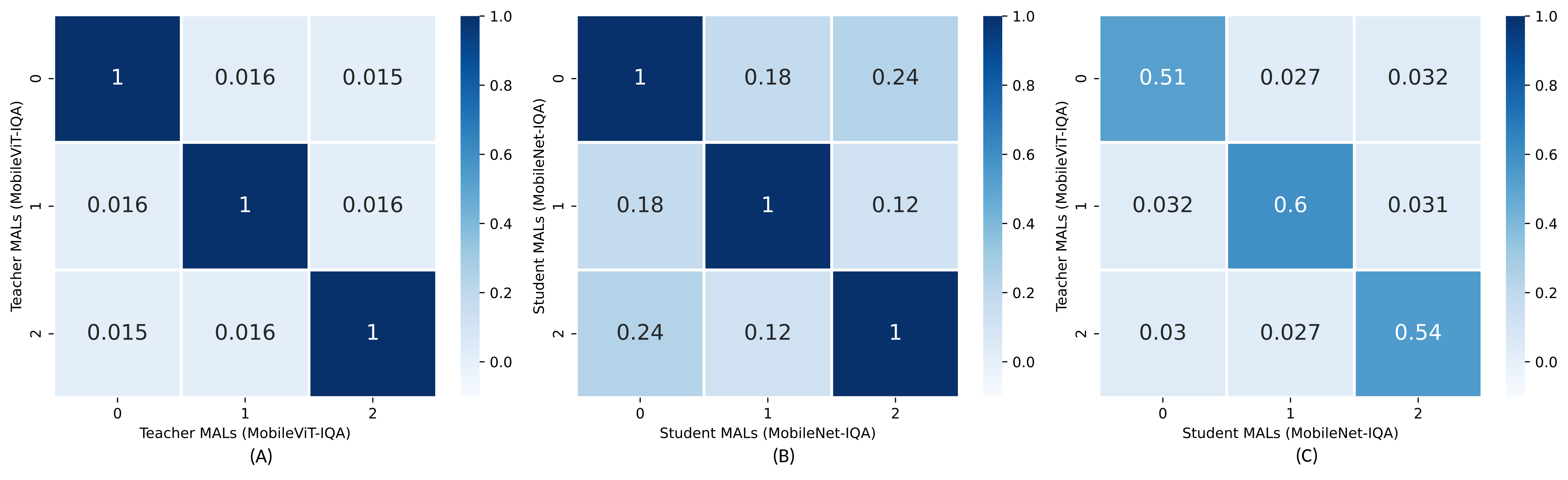}
  \caption{(A), (B) and (C) represent the cosine similarities of pairwise MALs within the MobileViT-IQA, MobileNet-IQA, and between MobileViT-IQA and MobileNet-IQA.}
  \label{fig:different_mals}
\end{figure}

To validate that the proposed MALs can learn diverse attentions, we compute the cosine similarity between the weights of each pairwise MALs and show it in \fig{fig:different_mals}-(A) and (B). We see that all the similarity scores except those in the diagonal are extremely low, meaning that there exists little redundancy between each pairwise MALs. Moreover, we compute the cosine similarity between the MALs of the teacher (MobileViT-IQA) and the student (MobileNet-IQA) to demonstrate whether the student learns from the teacher. As illustrated in \fig{fig:different_mals}-(C), the high diagonal similarity indicates that the distillation is effective at the corresponding positions, indicating that the student has successfully learned how to assess image from the teacher.

\begin{figure}[]
  \centering
\setlength{\abovecaptionskip}{0.1cm}
  \includegraphics[width=\columnwidth]{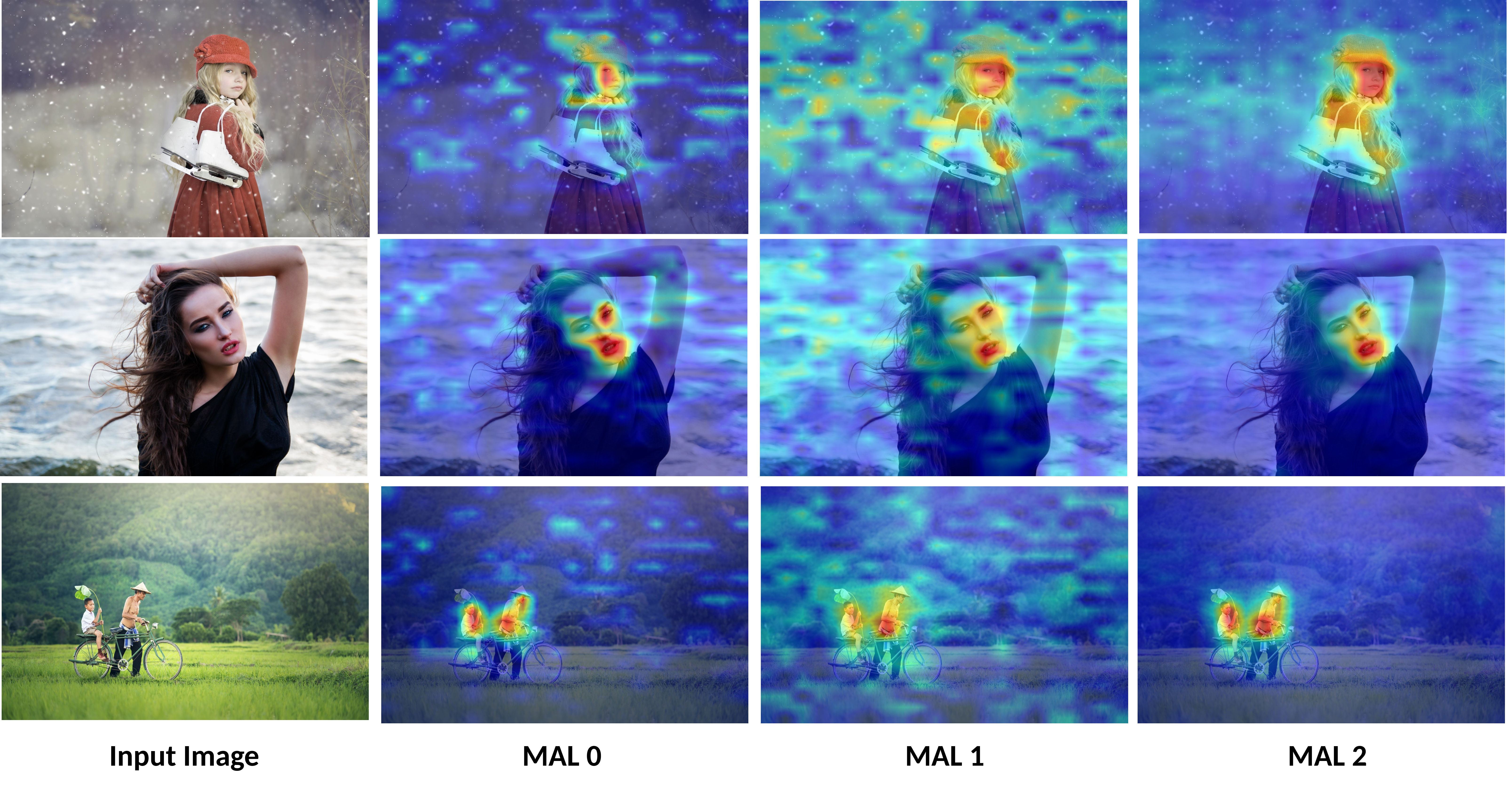}
  \caption{Attention maps produced by different MALs. The number of MALs is set to 3.
  }
  \label{fig:visual_mals}
\end{figure}

More intuitively, we visualize the output of different MALs in \fig{fig:visual_mals}. It can be observed that different MALs have distinct attention regions. For example, the first MAL pays more attention to local regions, the second and third MALs mainly focus on both global and local features. 
The examples show that each MAL effectively learns complementary opinion features.

\subsection{Running On Mobile Phones}

\begin{figure}[htbp]
  \centering
  \includegraphics[width=0.9\columnwidth]{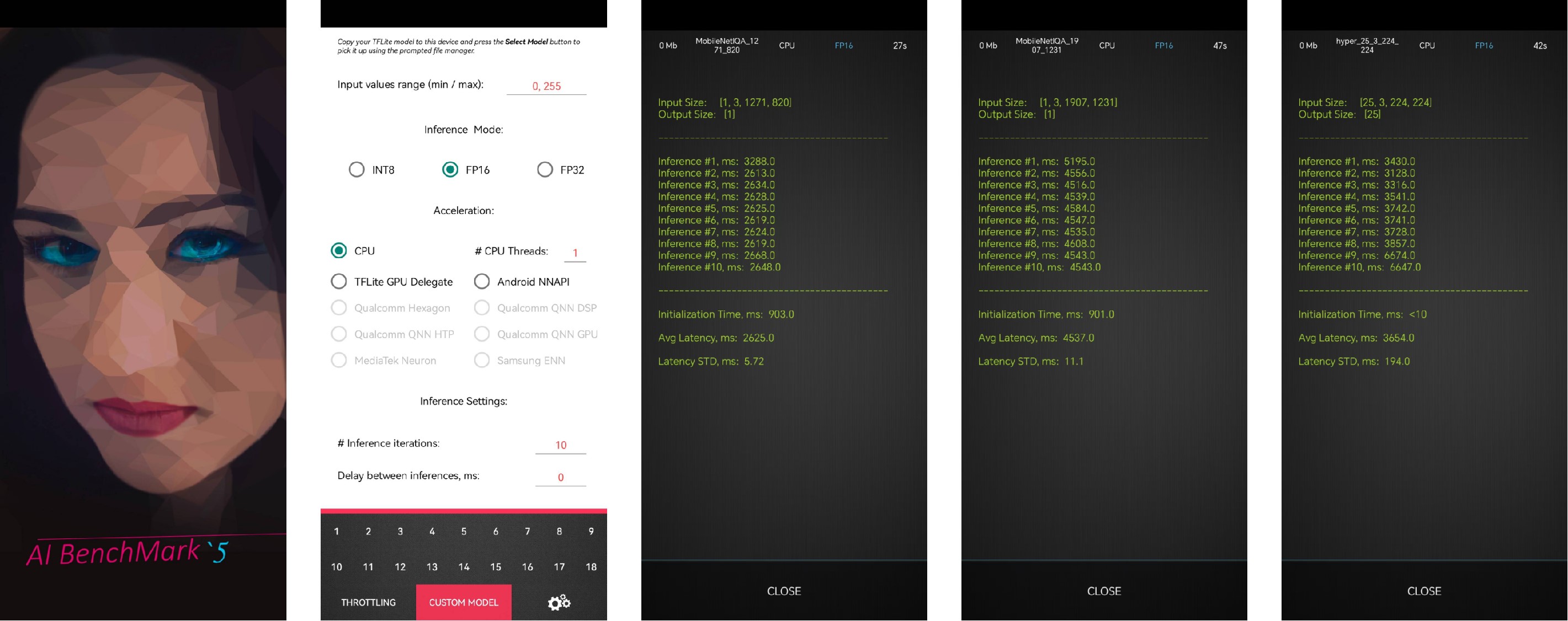}
  \caption{The AI Benchmark inference platform.}
  \label{fig:run}
\end{figure}

To validate the proposed MobileNet-IQA can be applied on the mobile devices, we convert the MobileNet-IQA and HyperIQA\cite{su2020blindly} into TensorFlow Lite (TFLite) and evaluate the inference efficiency on the AI Benchmark \footnote{https://ai-benchmark.com/}. We conduct the experiments on two mobile phones: Xiaomi 10S and HONOR Magic5 Pro. As illustrated in Fig. \ref{fig:run}, we set the inference mode to FP16 and run these models on a single CPU. This process is repeated 10 times, and the average of the 10 scores are reported as the final inference times (ms). The results shown in \tbl{tab:run_on_phone} demonstrate that MobileNet-IQA ($1271\times820$) not only shows faster model efficiency than HyperIQA, but also surpasses HyperIQA in overall model performance, further confirming the effectiveness of our approach.

\begin{table}[]
\caption{The inference time comparisons\protect\footnotemark   between MobileNetIQA and HyperIQA on different mobile phones. The model performance in terms of KRCC, PLCC and SRCC are provided for better comparison. The best restuls are marked in black bold.}
\label{tab:run_on_phone}
\centering
\scriptsize
\resizebox{\columnwidth}{!}{
\begin{tabular}{cc|cccc|cc}
\hline
\multirow{2}{*}{Model} & \multirow{2}{*}{Input Resolution} & \multirow{2}{*}{KRCC$\:\uparrow$} & \multirow{2}{*}{PLCC$\:\uparrow$} & \multirow{2}{*}{SRCC$\:\uparrow$} & \multirow{2}{*}{Average$\:\uparrow$} & \multicolumn{2}{c}{Inference Time (ms)} \\
 &  &  &  &  &  & Xiaomi 10S & HONOR Magic5 Pro \\ \hline
HyperIQA& $224 \times224$ & 0.359 & 0.182 & 0.524 & 0.355 & 5762 & 3654 \\ \hline
\multirow{6}{*}{MobileNetIQA} & $238\times 153$ & 0.316 & 0.477 & 0.458 & 0.417 & \textbf{1460} & \textbf{985} \\
 & $224 \times224$ & 0.339 & 0.505 & 0.488 & 0.444 & 1488 & 1003 \\
 & $317\times 205$ & 0.380 & 0.555 & 0.542 & 0.493 & 1522 & 1073 \\
 & $476\times 307$ & 0.456 & 0.652 & 0.637 & 0.582 & 1677 & 1145 \\
 & $1271\times 820 $ & 0.578 & 0.782 & 0.770 & 0.710 & 3636 & 2625 \\
 & $1907\times1231$ & \textbf{0.585} & \textbf{0.784} & \textbf{0.777} & \textbf{0.715} & 6465 & 4537 \\ \hline
\end{tabular}
}
\end{table}
\footnotetext{HyperIQA randomly crops $224\times224$ patches 25 times from the input image, and gets the quality score based on the average results of these 25 patches. MobileNet-IQA predict the quality score directly based on the full input image. In this experiment, the batch size (BS) for HyperIQA is 25, whereas the BS for MobileNet-IQA is 1.}

\subsection{Ablation Studies}
In this paper, we develop MobileIQA  based on the MAL module and employs knowledge distillation (KD) to train the student model (MobileNet-IQA) with the guidance from the teacher model (MobileViT-IQA).

To validate the effectiveness of these two key components, we conduct the following experiments. Firstly, we remove the three MALs in the MobileViT-IQA and re-train this model (W/O MAL). Then, we re-train the MobileNet-IQA directly without the guidance from the teacher model (W/O KD).
The results from \tbl{tab:ablation} reveal that the removal of any component degrades the model's performance. We can see that the variant removing the MAL (W/O MAL) has the most remarkable decline in performance, validating the significance of the diverse opinion feature learning. In addition, without the guidance from the teacher model, the W/O KD variant also shows a noticeable drop in performance. This indicates that the knowledge distillation effectively transfers reliable knowledge from the teacher model to the student model, enhancing the performance of the student model. Such a simple knowledge distillation approach can achieve this effect further validates the rationale behind our design of the diverse opinion network based on the MAL module.

\begin{table}[htbp]
\centering
\caption{Ablation studies on the critical components of our framework on the validation set. The average results of KRCC, PLCC and SRCC are provided. The best results are marked in black bold.}
\label{tab:ablation}
\begin{tabular}{cc|cc|cccc}
\toprule
Model & Variant & RMSE$\:\downarrow$ & MAE$\:\downarrow$ & KRCC$\:\uparrow$ & PLCC$\:\uparrow$ & SRCC$\:\uparrow$ &  Average$\:\uparrow$ \\ \midrule
MobileViT-IQA & W/O MAL & 0.046 &0.036 	&0.556 	&0.750 	&0.748 	&0.685  \\ \cline{2-8} 
(Teacher)  & Full & \textbf{0.043} & \textbf{0.034} & \textbf{0.585} & \textbf{0.784} & \textbf{0.777} & \textbf{0.715} \\ \midrule
MobileNet-IQA  & W/O KD & 0.045 & 0.035 & 0.562 & 0.759 & 0.754 & 0.692 \\ \cline{2-8} 
(Student)& Full & \textbf{0.044} & \textbf{0.034} & \textbf{0.582} & \textbf{0.783} & \textbf{0.755} & \textbf{0.707} \\\bottomrule
\end{tabular}
\end{table}

\section{Conclusion}
In this paper, we introduce MobileIQA, an innovative framework comprising a powerful teacher model (MobileViT-IQA) and a lightweight student model (MobileNet-IQA). Both models leverage lightweight networks, MobileViT and MobileNet, as their backbones, respectively. We significantly increase the input resolution from the $224\times224$ to $1907\times1231$, enhancing model performance by capturing more image detail. Furthermore, both models incorporate our proposed Multi-view Attention Learning modules, which provide diverse perspectives on input images and enhance network performance. The student model is trained with the guidance of the teacher model, achieving strong performance with much smaller computational complexity. Extensive experiments demonstrate the superior accuracy and computational efficiency of our approach.

\section*{Acknowledgements}

\parbox{\textwidth}{
This work was partially supported by the Humboldt Foundation. We thank the AIM 2024 sponsors: Meta Reality Labs, KuaiShou, Huawei, Sony Interactive Entertainment and University of W\"urzburg (Computer Vision Lab).
Additionally, this work is also supported by the Key Research and Development Program of Xinjiang Urumgi Autonomous Region under Grant No.2023B01005, the Natural Science Foundation of China (Nos.62122086), the Natural Science Foundation of China under Grants 62202470. Bing Li is also supported by Youth Innovation Promotion Association, CAS.
}

\bibliographystyle{splncs04}
\bibliography{main}
\end{document}